\pdfoutput=1
\documentclass[journal]{IEEEtran}
\IEEEoverridecommandlockouts

\usepackage{amsmath,amssymb,amsfonts}
\usepackage{graphicx}
\usepackage[ruled,vlined,linesnumbered,noend]{algorithm2e}
\usepackage{hyperref}
\usepackage{subcaption}
\hypersetup{hidelinks=true}
\usepackage{textcomp}
\usepackage{booktabs}
\usepackage{makecell}
\usepackage{amsthm}
\usepackage{tikz}
\usetikzlibrary{positioning}
\usepackage{xspace}
\usepackage[most]{tcolorbox}
\usepackage{multirow}

\definecolor{queryblue}{HTML}{E3F2FD}
\definecolor{modelgreen}{HTML}{E8F5E9}
\definecolor{baseorange}{HTML}{FFF3E0}
\definecolor{refgray}{HTML}{F5F5F5}

\makeatletter
\patchcmd{\@algocf@start}{-1.5em}{0pt}{}{}
\makeatother

\SetCommentSty{mycomment}

\def\BibTeX{{\rm B\kern-.05em{\sc i\kern-.025em b}\kern-.08em
    T\kern-.1667em\lower.7ex\hbox{E}\kern-.125emX}}

\newcommand\sysname{PrivMedChat\xspace}

\newcommand{\codelink}{https://github.com/sudip-bhujel/privmedchat}

\newcommand*\circled[1]{\tikz[baseline=(char.base)]{\node[shape=circle,draw,inner sep=0.6pt] (char) {\small #1};}}

\newtheorem{definition}{Definition}

\begin{document}
\title{
  \sysname: End-to-End Differentially Private RLHF for Medical Dialogue Systems

}
\author{

  Sudip Bhujel
  \thanks{Sudip is a Graduate Student in the Department of Computer Science, University of Kentucky, Lexington, KY, USA (email: \href{mailto:sudipbhujel@uky.edu}{sudipbhujel@uky.edu})}
}

\maketitle

\begin{abstract}

    Large language models are increasingly used for patient-facing medical assistance and clinical decision support, but adapting them to clinical dialogue often requires supervision derived from doctor-patient conversations that may contain sensitive information. Conventional supervised fine-tuning and reinforcement learning from human feedback (RLHF) can amplify memorization, enabling membership inference and disclosure of rare training-set details. We present \sysname (\textbf{Priv}ate \textbf{Med}ical \textbf{Chat}), an end-to-end framework for differentially private RLHF (DP-RLHF) for medical dialogue systems. Our approach enforces differential privacy at each training stage that accesses dialogue-derived supervision, combining DP-SGD for supervised fine-tuning and reward model learning from preference pairs, and DP-aware policy optimization for alignment. To avoid costly clinician labeling, we introduce an annotation-free preference construction strategy that pairs physician responses with filtered non-expert generations. We evaluate \sysname across medical dialogue tasks and assess utility, safety, and privacy under consistent privacy accounting, thereby providing a practical pathway to align medical chatbots while offering formal privacy guarantees. We open-source our code at \url{\codelink}.
\end{abstract}


\section{Introduction}
\label{sec:intro}

Large Language Models (LLMs)
are increasingly used for clinical decision support and patient-facing assistance, ranging from diagnosis and symptoms to care navigation and patient education. Recent medical-domain systems such as Med-PaLM~\cite{singhal2025medpalm}, ChatDoctor~\cite{li2023chatdoctor}, and BioMixtral~\cite{labrak2024biomistral} demonstrate that adapting general-purpose LLMs to biomedical knowledge and clinical dialogue can improve response helpfulness, domain fluency, and adherence to medical communication norms. These capabilities make medical chatbots promising tools for scaling access to health information and reducing clinician burden, especially in resource-constrained settings.

\begin{figure}
  \centering
  \includegraphics[width=0.6\linewidth]{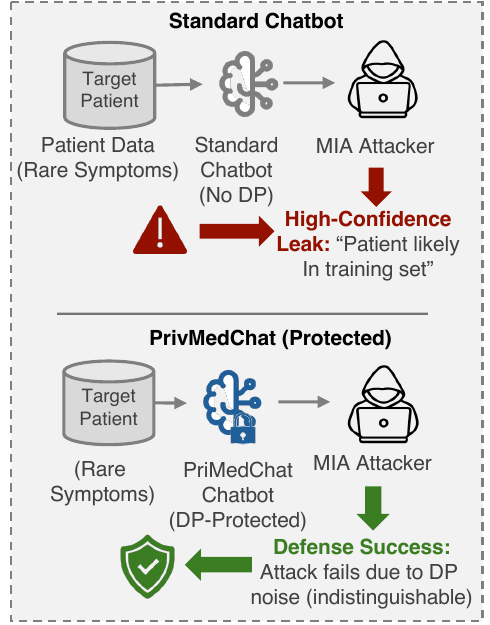}
  \caption{A standard medical LLM that is fine-tuned without privacy safeguards may disclose the membership for patients with rare symptoms in its training set. In contrast, \sysname's application of differential privacy prevents high-confidence inferences.}
  \label{fig:threat-model}
\end{figure}

State-of-the-art alignment of such systems commonly follows a Reinforcement Learning from Human Feedback (RLHF)~\cite{ouyang2022training} pipeline. Training a medical LLM entirely from scratch is both time-consuming and ineffective; instead, it is common practice to fine-tune a strong pre-trained model dialogue LLM. The process begins by adapting the base model through Supervised Fine-Tuning (SFT) on high-quality clinical interactions to learn domain style and conversational structure. Next, a reward model is trained on ranked preference pairs to approximate expert judgments of clinical utility and safety. Finally, the SFT model is optimized to maximize reward, typically via Proximal Policy Optimization (PPO)~\cite{schulman2017proximal} or related preference-optimization methods. This preference-based stage is often critical for improving safety behavior (e.g., appropriate escalation) and for reducing unhelpful or overly generic responses.

Despite these advances, effective medical adaptation and personalization often rely on training signals derived from doctor-patient dialogues, which can contain protected health information (PHI) and other sensitive attributes governed by regulations such as the Health Insurance Portability and Accountability Act (HIPAA) and the General Data Protection Regulation (GDPR). Conventional fine-tuning and RLHF may inadvertently memorize training records, enabling membership inference attacks (MIA)~\cite{duan2024membership}, attribute inference, or verbatim leakage of rare patient details~\cite{carlini2021extracting}. Importantly, even datasets released in de-identified or ``public'' form can remain sensitive in an adversarial setting: rare strings, uncommon symptom combinations, or residual identifiers may still be memorized and later recoverable. In high-stakes clinical contexts, such leakage risks patient re-identification and undermines trust in deployment.

Differential privacy (DP) provides a rigorous framework for limiting the influence of any single training record on model parameters~\cite{dwork2014algorithmic}. Prior work has successfully applied DP to language model pre-training and supervised fine-tuning (DP-SFT)~\cite{abadi2016deep,yu2021differentially,li2024fine}, showing that nontrivial utility can be retained under meaningful privacy budgets. However, extending DP to the full RLHF pipeline remains challenging. Both preference modeling and policy optimization are often highly sensitive to noise, and naive DP mechanisms can significantly degrade alignment quality---an issue that is especially critical in healthcare, where correctness, safety, and calibrated uncertainty are essential. Consequently, existing privacy-preserving LLMs frequently exhibit pronounced performance drops at privacy levels that meaningfully reduce empirical leakage risk~\cite{singarajah2023benchmarking}.

This paper addresses the challenge of extending DP beyond supervised fine-tuning to the full preference-based alignment pipeline for medical dialogue. We propose \sysname, an end-to-end differentially private RLHF framework that applies DP-SGD~\cite{abadi2016deep} to (i) SFT on dialogue-derived supervision, (ii) reward model training on preference pairs, and (iii) PPO policy optimization. Because the PPO stage also accesses dialogue-derived prompts, we apply DP-SGD to both the PPO actor and critic, accounting for their privacy cost in $\varepsilon_{\text{total}}$. The DP-trained reward model is kept fixed during alignment so that its privacy budget is not further consumed.

Although some medical dialogue datasets are publicly released in de-identified form, we treat dialogue-derived supervision as \emph{sensitive} due to residual identifiers, rare symptom combinations, and adversarial linkage risks. The DP guarantees reported here apply to the training records used for SFT and reward modeling and limit the influence of any single example on the released model parameters, not as a claim of regulatory compliance.

To avoid costly clinician annotation, we construct preference pairs using an expert-versus-non-expert strategy from medical dialogue corpora. Specifically, we treat real physician responses as ``chosen'' outputs and generate ``rejected'' alternatives using a base LLM prompted to respond as a non-expert assistant; we apply similarity and quality filtering to ensure informative preference margins. This produces scalable, high-signal preference data for reward modeling without requiring additional human labeling.

Our primary contributions are as follows:
\begin{enumerate}
  \item \textbf{Annotation-free medical preference construction.} We present a scalable strategy to form preference pairs for medical dialogue alignment by pairing physician responses with filtered non-expert generations, producing reward-model training data without clinician labeling.
  \item \textbf{End-to-end DP-RLHF.} We introduce a DP-protected training pipeline that applies DP-SGD~\cite{abadi2016deep} to all three stages---SFT, reward modeling, and PPO---ensuring formal $(\varepsilon,\delta)$ privacy guarantees across the full alignment pipeline.
  \item \textbf{Utility--safety--privacy evaluation under DP.} We evaluate utility with automatic metrics, safety with targeted heuristic checks, and privacy with membership inference and canary-style probing, and we report results across multiple privacy budgets using consistent accounting across training stages.
\end{enumerate}

\begin{table}
  \centering
  \caption{List of Notations}
  \label{tab:notation}
  \footnotesize
  \renewcommand{\arraystretch}{1.05}
  \begin{tabular}{ll}
    \toprule
    \textbf{Symbol}                 & \textbf{Meaning}               \\
    \midrule
    $D_{\text{priv}}^{\text{SFT}}$  & Private SFT corpus             \\
    $D_{\text{priv}}^{\text{pref}}$ & Private preference pairs       \\
    $\mathcal{X}_{\text{pub}}$      & Public/synthetic prompts (PPO) \\
    $\varepsilon_{\text{SFT}}$         & Privacy cost of DP-SFT         \\
    $\varepsilon_{\text{RM}}$          & Privacy cost of reward model   \\
    $\varepsilon_{\text{total}}$       & Total privacy cost (composed)  \\
    $C$                             & DP-SGD clip norm               \\
    $\sigma$                        & DP-SGD noise multiplier        \\
    $r_\phi(x,y)$                   & Reward model score             \\
    $\pi_{\text{ref}}$              & Reference policy (SFT)         \\
    $\beta$                         & KL penalty weight              \\
    $(x,y_w,y_l)$                   & Preference triple (win/lose)   \\
    $\varepsilon_{\text{clip}}$        & PPO clipping parameter         \\
    \bottomrule
  \end{tabular}
\end{table}

\noindent\textbf{Organization.} The remainder of this paper is organized as follows. Section~\ref{sec:methods} outlines the methodology and the \sysname training pipeline. Section~\ref{sec:implementation} describes the implementation, and Section~\ref{subsec:evaluation} presents the experimental evaluation, including analyses of utility, safety, and privacy. Section~\ref{sec:related} reviews related work on medical LLMs, alignment, and privacy. Section~\ref{sec:discussion} discusses limitations and ethical considerations, and Section~\ref{sec:conclusion} concludes the paper. Appendix~\ref{sec:prelim} presents the preliminaries on differential privacy, efficient fine-tuning techniques, and RLHF. Appendix~\ref{app:dp-accounting} describes the differential privacy accounting of the system, and Appendix~\ref{app:qualitative-examples} compares the qualitative results of the base model (without fine-tuning), physician responses, and \sysname (fine-tuned with privacy).

\section{Methods}
\label{sec:methods}

\begin{figure*}[htbp]
  \centering
  \begin{subfigure}[b]{0.40\textwidth}
    \centering
    \includegraphics[height=5.2cm, keepaspectratio]{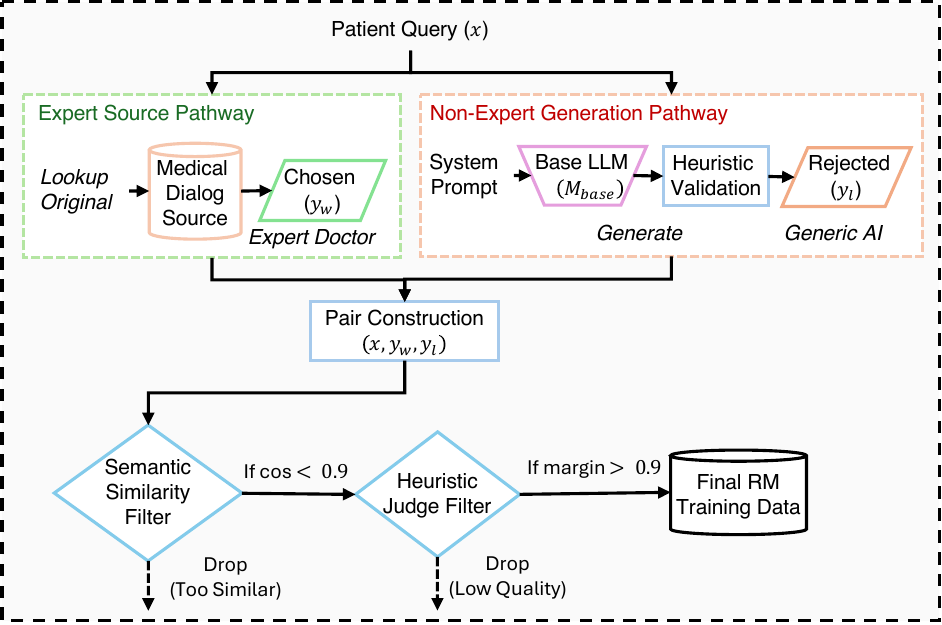}
    \caption{Pre-processing of data}
    \label{fig:preprocess}
  \end{subfigure}
  \hfill
  \begin{subfigure}[b]{0.58\textwidth}
    \centering
    \includegraphics[height=5.2cm, keepaspectratio]{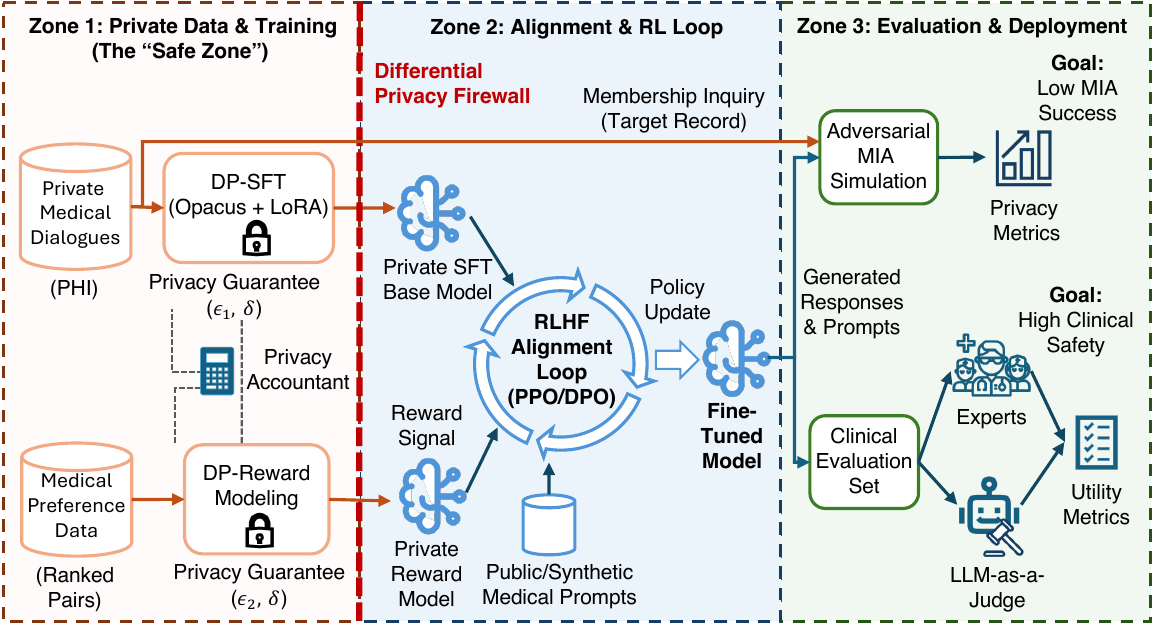}
    \caption{System architecture of \sysname}
    \label{fig:sys}
  \end{subfigure}
  \caption{Data pre-processing and system overview of \sysname.
    (a) methodology for annotation-free medical preference pair construction
    (b) illustrates the three-zone design separating DP-protected training (Zone 1) from evaluation and deployment (Zones 2--3).}

  \label{fig:threat-sys}
\end{figure*}

We propose \sysname, a differentially private framework for medical dialogue generation and alignment that enforces differential privacy across the RLHF pipeline. Figure~\ref{fig:sys} summarizes the architecture, organized into three zones. Zone~\circled{1} contains the private dataset and all DP-protected training procedures (DP-SFT, DP reward modeling, and DP-PPO); Zone~\circled{2} handles evaluation and deployment; and Zone~\circled{3} assesses both privacy and downstream utility. The overall procedure is summarized in Algorithm~\ref{alg:dprlhf}, and Table~\ref{tab:notation} lists the notations used throughout the paper.

\begin{algorithm}[htbp]
  \caption{End-to-End DP-RLHF (\sysname)}
  \label{alg:dprlhf}
  \KwIn{Sensitive corpus $D_{\text{priv}}$, Base model $\pi_\text{base}$, LLM Evaluator, Targets $(\varepsilon, \delta)$}
  \KwOut{Privacy-aligned medical LLM policy $\pi_\theta$}
  \BlankLine
  \tcc{Stage 1: Preference Pair Construction (Zone 1)}
  Initialize $D_{\text{pref}} \gets \emptyset$\;
  \ForEach{$x_i \in D_{\text{priv}}$}{
    Sample candidate responses from $\pi_\text{base}(\cdot | x_i)$\;
    Score candidates via LLM Evaluator to find $(y_i^w, y_i^l)$\;
    $D_{\text{pref}} \gets D_{\text{pref}} \cup \{(x_i, y_i^w, y_i^l)\}$\;
  }
  \BlankLine
  \tcc{Stage 2: DP-SFT (Zone 1)}
  Initialize $\theta_\text{sft}$ from $\pi_\text{base}$\;
  \For{$t=1$ \KwTo $T_\text{sft}$}{
    Sample batch $B \sim D_{\text{priv}}$\;
    Compute per-sample gradients $g_i = \nabla \mathcal{L}_\text{SFT}^{(i)}(\theta_\text{sft})$\;
    Clip and aggregate: $g_t = \frac{1}{|B|}\sum_{i \in B} g_i / \max\left(1, \frac{\|g_i\|_2}{C_\text{sft}}\right)$\;
    Add noise: $\tilde{g}_t = g_t + \mathcal{N}(0, \sigma_\text{sft}^2 C_\text{sft}^2 \mathbf{I})$\;
    Update $\theta_\text{sft}$ using $\tilde{g}_t$\;
  }
  \BlankLine
  \tcc{Stage 3: DP-RM (Zone 1)}
  Initialize reward model $\phi$ from $\theta_\text{sft}$\;
  \For{$t=1$ \KwTo $T_\text{rm}$}{
    Sample batch $B \sim D_{\text{pref}}$\;
    Compute per-sample gradients $g_i = \nabla \mathcal{L}_\text{RM}^{(i)}(\phi)$\;
    Clip and aggregate: $g_t = \frac{1}{|B|}\sum_{i \in B} g_i / \max\left(1, \frac{\|g_i\|_2}{C_\text{rm}}\right)$\;
    Add noise: $\tilde{g}_t = g_t + \mathcal{N}(0, \sigma_\text{rm}^2 C_\text{rm}^2 \mathbf{I})$\;
    Update $\phi$ using $\tilde{g}_t$\;
  }
  \BlankLine
  \tcc{Stage 4: DP-PPO Policy Optimization}
  Initialize $\pi_\theta$ from $\theta_\text{sft}$\;
  Sample dialogue-derived prompts $D_\text{ppo}$\;
  \For{$t=1$ \KwTo $T_\text{ppo}$}{
    Sample prompts $x \sim D_\text{ppo}$\;
    Generate responses $y \sim \pi_\theta(\cdot | x)$\;
    Evaluate rewards $R(x, y) = r_\phi(x, y) - \beta \mathbb{D}_\text{KL}[\pi_\theta \| \pi_\text{sft}]$\;
    Update $\pi_\theta$ via PPO with DP-SGD on actor and critic\;
  }
  \Return{$\pi_\theta$}\;
\end{algorithm}

\vspace{0.01in}
Let $D_{\text{priv}}^{\text{SFT}}$ denote the (potentially sensitive) supervised fine-tuning corpus derived from doctor--patient conversations, and let $D_{\text{priv}}^{\text{pref}}=\{(x,y_w,y_l)\}$ denote the preference-pair dataset used for reward modeling, where $y_w$ is the physician (chosen) response and $y_l$ is the rejected response. Let $\mathcal{X}_{\text{pub}}$ denote a set of public or synthetic prompts that contain no real patient information. In \sysname, DP guarantees are enforced for training procedures that access $D_{\text{priv}}^{\text{SFT}}$ or $D_{\text{priv}}^{\text{pref}}$ (Zone~\circled{1}), while policy alignment uses only $\mathcal{X}_{\text{pub}}$ together with a fixed DP-trained reward model (Zone~\circled{2}).

We initialize our SFT and RM from \texttt{Meta-Llama-3-8B-Instruct} and use it as the starting checkpoint for all subsequent fine-tuning; we refer to this checkpoint as the \emph{base model} hereafter.

\subsection{Dataset and Preprocessing}
\label{subsec:data}

We utilize the medical dialogue corpus \texttt{OpenMed/MedDialog}~\cite{openmed_meddialog_2026}, which has been de-identified using Presidio~\cite{mendels2018presidio}. This de-identification process replaces sensitive information with placeholders, including the removal of email and phone number patterns and signature-like text. The DP guarantees are defined with respect to individual training examples derived from these dialogues. The raw dialogues are transformed into structured input–output pairs for SFT and preference learning. Additionally, we apply lightweight text normalization and implement group-aware dataset splitting based on conversation identifiers to prevent cross-split leakage.

To support reward modeling without human annotation, we generate ``rejected'' responses using a general-purpose language model (\texttt{meta-llama/Llama-3.1-8B}) prompted to simulate a non-expert medical assistant. We then apply heuristic quality filtering to remove low-quality samples, discarding outputs shorter than 10 words, containing explicit refusal patterns (e.g., ``I am an AI''), or exhibiting substantial repetition. Because reward modeling benefits from a clear margin between preferred and dispreferred outputs, we further deduplicate preference pairs by semantic similarity. Specifically, each response is embedded using a sentence transformer (\texttt{all-MiniLM-L6-v2}), and we remove pairs whose chosen (expert) and rejected (generated) responses exceed a cosine similarity threshold of $0.90$.

Finally, we partition the dataset into training, validation, and test splits using a group-aware strategy keyed by source conversation identifiers, ensuring that turns from the same multi-turn session do not appear across multiple splits.
The resulting corpus yields high-quality, de-duplicated supervision and preference pairs that clearly separate expert medical advice from generic assistant responses.

\subsection{Preference Pair Construction}
\label{subsec:pref}

Reward model training relies on ranked preference pairs $(x, y_w, y_l)$ constructed via an expert-versus-non-expert strategy that avoids manual annotation. The chosen response $y_w$ is the original physician response provided in the dataset, while the rejected response $y_l$ is generated by prompting a general-purpose generator model (\texttt{meta-llama/Llama-3.1-8B}) to behave as a non-expert assistant. Specifically, the generative mode is conditioned on the patient query along with a system instruction to provide concise, high-level guidance while explicitly avoiding detailed differential diagnosis. The prompt takes the following form:

{
\small
\begin{verbatim}
[System]: ... Avoid detailed differential
          diagnosis ...
[Patient]: {patient_text}
[Doctor]:
\end{verbatim}
}

We use nucleus sampling ($p=0.95$) at a temperature $0.7$, and we restrict outputs to matched lengths with a maximum of 256 tokens. This constraint ensures that rejected responses remain plausible while being consistently less specific than the expert reference.

To ensure the constructed pairs contain a meaningful preference signal, we apply a multi-stage filtering pipeline. First, we remove degenerate generations using heuristic criteria, including minimum length requirements, refusal-pattern detection, and repetition filtering. Next, we perform semantic filtering by embedding both the physician response and the generated response with \texttt{sentence-transformers/all-MiniLM-L6-v2} and discarding any pairs whose cosine similarity is $\ge 0.90$. Optionally, we apply a heuristic judge that computes a composite score incorporating length penalties, medical terminology density, and safety-refusal detection, retaining only those pairs in which the chosen response surpasses the rejected response by at least $0.20$. In practice, this procedure produces approximately $10\text{k}$--$50\text{k}$ preference pairs and encourages the reward model to prefer clinically grounded, expert-level responses over generic or hedged alternatives.

\subsection{Supervised Fine-Tuning}
\label{subsec:sft}

We adapt the \emph{base model} to the medical dialogue domain via supervised fine-tuning on the processed conversations. Because these data may contain sensitive patient information, we perform DP-SFT using DP-SGD~\cite{abadi2016deep}. 
Our implementation uses Opacus~\cite{yousefpour2021opacus} for privacy accounting and the Parameter Efficient Fine Tuning (PEFT) technique for training. 
We freeze the backbone model parameters and train only Low Rank Adaptation (LoRA)~\cite{hu2022lora} adapters inserted into the query, key, value, and output projection modules.

\vspace{0.25em}
\textbf{Privacy Guarantees.} We enforce per-example gradient norm bounds using DP-SGD~\cite{abadi2016deep} with ``ghost clipping,'' constraining the $\ell_2$ norm of each per-sample gradient to at most $1.0$. We then add Gaussian noise to the aggregated gradients to achieve a strict privacy target budget, tracking privacy loss using a R\'enyi Differential Privacy (RDP)~\cite{mironov2017renyi} accountant. This mechanism limits the influence of any single training example on the learned parameters, reducing the likelihood that the model memorizes identifiable patient content while still learning general medical dialogue patterns.

\subsection{Differentially Private RLHF}
\label{subsec:dprlhf}

Preference-based alignment is performed in two DP-protected phases within the secure environment of Zone~\circled{1}, providing end-to-end privacy with respect to the sensitive training corpus.

\subsubsection{DP-Reward Modeling}
\label{subsubsec:rm}

We train a reward model by fine-tuning a separate copy of the \emph{base model} on the constructed preference pairs using the same DP-SGD~\cite{abadi2016deep} stack (Opacus~\cite{yousefpour2021opacus} + LoRA~\cite{hu2022lora}) employed in DP-SFT. The reward model learns to assign higher scalar scores to expert physician responses $y_w$ than to non-expert generations $y_l$. Concretely, we optimize a standard pairwise preference loss (Bradley--Terry / logistic form):
\[
  \mathcal{L}_{\text{RM}}(\phi)=
  -\mathbb{E}_{(x,y_w,y_l)}\left[
    \log \sigma\left(r_\phi(x,y_w)-r_\phi(x,y_l)\right)
    \right],
\]
where $r_\phi(x,y)$ is the scalar reward assigned to response $y$ under prompt $x$ and $\sigma(\cdot)$ is the sigmoid. The privacy expenditure for this stage, denoted $\varepsilon_{\mathrm{RM}}$, is accounted for independently from SFT.

\subsubsection{Policy Optimization with Firewall}
\label{subsubsec:ppo-firewall}

We align the DP-SFT policy using PPO~\cite{schulman2017proximal}. We initialize the PPO actor from the DP-SFT checkpoint and use the same DP-SFT checkpoint as the reference policy $\pi_{\mathrm{ref}}$ for KL regularization. The private reward model---trained under DP in the previous phase---provides the optimization signal. We additionally regularize the policy to remain close to a reference model $\pi_{\text{ref}}$ (typically the SFT model) via an explicit KL penalty:

\vspace{-0.15in}
\begin{equation*}
  \footnotesize
  \label{eq:kl_obj}
  \begin{aligned}
    \max_{\theta}\quad
    \mathbb{E}_{x,\, y\sim \pi_\theta(\cdot\mid x)}
    \big[r_\phi(x,y)\big]
    -\beta\,
    \mathbb{E}_{x}
    \Big[\mathrm{KL}\Big(\pi_\theta(\cdot\mid x)\,\big\| \pi_{\mathrm{ref}}(\cdot\mid x)\Big)\Big].
  \end{aligned}
\end{equation*}

Because the PPO prompts are derived from the same dialogue corpus used for SFT and reward modeling, this stage also accesses dialogue-derived data. To maintain end-to-end privacy, we therefore apply DP-SGD~\cite{abadi2016deep} to both the PPO actor and critic, with separate per-component privacy budgets.
The total privacy cost of the system is thus the composition of all three stages: $\varepsilon_{\text{total}} = \varepsilon_{\mathrm{SFT}} + \varepsilon_{\mathrm{RM}} + \varepsilon_{\mathrm{PPO}}$, with $\delta$ tracked analogously using the same RDP accountant~\cite{mironov2017renyi}.

\subsection{Evaluation and Deployment}
\label{subsec:eval-deploy}

Zone~\circled{3} (Evaluation and Deployment) operates outside the DP-protected training pipeline and accesses no private patient data or preference pairs. Evaluations are performed on held-out test prompts and generated responses, while deployment uses only user-provided inputs at inference time.

\subsubsection{Evaluation}

We evaluate \sysname along three axes: utility, safety, and privacy. Automatic utility metrics (ROUGE-L~\cite{lin2004rouge}, BERTScore~\cite{zhang2019bertscore}, entity overlap, and perplexity) are computed on a larger held-out subset (5,000 prompts) to reduce variance in aggregate reporting. For safety checks and privacy attacks, we use a focused evaluation subset of $n=500$ prompts and report 1,000-iteration paired bootstrap confidence intervals. All evaluation prompts are drawn from held-out test data and are not used during training.

No private corpus is touched during these assessments; all prompts are non-sensitive.

\subsubsection{Deployment Considerations}

In deployment, \sysname processes only real-time user prompts, inheriting the formal $(\varepsilon,\delta)$-DP guarantees from the DP-protected training stages. Practical safeguards include real-time monitoring for out-of-distribution inputs, human-in-the-loop escalation for high-risk queries, audit logging, and uncertainty-aware refusal mechanisms. The DP-SGD~\cite{abadi2016deep} framework applied across all training stages provides formal privacy-by-design guarantees. We do not frame DP guarantees as a standalone compliance claim; real deployments still require institutional governance, monitoring, and clinical safety processes.

\section{Implementation}
\label{sec:implementation}

We implement \sysname in Python. The codebase supports DP-SFT, DP reward modeling, and PPO-based RLHF with LoRA~\cite{hu2022lora} and is publicly available~\footnote{\url{\codelink}}. Differential privacy is enforced in all training stages that utilize dialogue-derived supervision through DP-SGD~\cite{abadi2016deep}, implemented with Opacus~\cite{yousefpour2021opacus}. We train \sysname{} on 1--4 NVIDIA H200 GPUs.

All components are initialized from the \emph{base model} to leverage its dialogue and reasoning capabilities. DP training uses Opacus~\cite{yousefpour2021opacus} with per-sample gradients and ghost clipping'' to constrain $\ell_2$ norms. SFT and reward modeling apply the same DP-SGD~\cite{abadi2016deep} to LoRA~\cite{hu2022lora} parameters with the same RDP accountant~\cite{mironov2017renyi}.

During DP-SFT, adapters are inserted into the query, key, value, and output projection layers with $r = 16$, $\alpha = 32$, and a dropout rate of $0.1$. The reward model employs adapters in the query and value projections with $r = 8$, $\alpha = 16$, and dropout $0.2$. The PPO~\cite{schulman2017proximal} actor and critic use adapters in the query and value projections with $r = 8$, $\alpha = 32$, and dropout $0.1$.

\vspace{0.25em}
\textbf{Training hyperparameters and DP accounting.}
Hyperparameters are chosen to maintain stable optimization under per-sample clipping and noise injection. We adopt a staged learning-rate schedule, using a conservative learning rate during PPO~\cite{schulman2017proximal} to mitigate policy collapse when combining RLHF with DP noise. Noise multipliers for DP-SGD~\cite{abadi2016deep} are computed with an RDP-based~\cite{mironov2017renyi} accountant to ensure that cumulative privacy loss remains within the target $\varepsilon$ budget across all three private training stages (DP-SFT, DP reward modeling, and DP-PPO). Table~\ref{tab:hyperparameters} summarizes the principal settings.

\begin{table}[htbp]  
  \centering
  \caption{Hyperparameter configurations for the \sysname training stages. Parameters were selected to balance convergence stability with differential privacy noise requirements.}
  \label{tab:hyperparameters}
  \begin{tabular}{@{}lccc@{}}
    \toprule
    \textbf{Parameter}   & \textbf{SFT}       & \textbf{Reward Model} & \textbf{\makecell[c]{PPO~\cite{schulman2017proximal} \\(Actor/Critic)}} \\
    \midrule
    Effective Batch Size & 16 -- 64           & 16 -- 64              & 16 -- 64                                             \\
    Learning Rate        & $2 \times 10^{-5}$ & $5 \times 10^{-5}$    & $5 \times 10^{-6}$                                   \\
    Training Duration    & 3 Epochs           & 3 Epochs              & 10 Iterations                                        \\
    Epochs per Iteration & --                 & --                    & 3                                                    \\
    Optimization Steps   & --                 & --                    & $\approx 1{,}000$                                    \\
    \bottomrule
  \end{tabular}
\end{table}

\begin{table*}[t]
  \centering
    \setlength{\tabcolsep}{2.5pt} %
  \renewcommand{\arraystretch}{0.95} %
  \caption{Utility and privacy results on the held-out test set. \textbf{Bold} denotes the best result per metric among DP models. $\varepsilon$ denotes the cumulative privacy budget ($\delta{=}10^{-5}$). Utility metrics (ROUGE-L~\cite{lin2004rouge}, BERTScore~\cite{zhang2019bertscore}, Entity F1, PPL) are computed on 5{,}000 prompts. PubMedQA~\cite{jin2019pubmedqa} accuracy is evaluated on 500 biomedical questions. Privacy columns report AUC-ROC of six membership inference attacks; values near $0.50$ indicate random-guessing (ideal).}
  \label{tab:results}
  \resizebox{\textwidth}{!}{%
    \begin{tabular}{llccccccccccc}
      \toprule
                     &                         & \multicolumn{5}{c}{\textbf{Utility}} & \multicolumn{6}{c}{\textbf{Privacy (MIA AUC $\downarrow$)}}                                                                                                                                                                                                                                                                                                                                      \\
      \cmidrule(lr){3-7}\cmidrule(lr){8-13}
      \textbf{Model} & $\boldsymbol{\varepsilon}$ & \textbf{ROUGE-L}                     & \textbf{BERTScore}                                          & \textbf{Entity F1} & \textbf{PPL $\downarrow$} & \textbf{PubMedQA}~\cite{jin2019pubmedqa} & \textbf{Loss}~\cite{yeom2018privacy} & \textbf{Ref} & \textbf{Min-K\%}~\cite{shi2023detecting} & \textbf{Min-K\%++}~\cite{zhang2024min} & \textbf{Zlib}~\cite{carlini2021extracting} & \textbf{Lowercase}~\cite{carlini2021extracting} \\
      \midrule
      Base           & --                      & 0.133                                & 0.830                                                       & 0.112              & 37.34                     & 0.554                                    & 0.518                                & 0.500        & 0.513                                    & 0.505                                  & 0.513                                      & 0.502                                           \\
      SFT            & --                      & 0.163                                & 0.829                                                       & 0.096              & 19.95                     & \textbf{0.642}                           & 0.532                                & 0.542        & 0.535                                    & 0.548                                  & 0.532                                      & 0.515                                           \\
      Overfit SFT    & --                      & 0.162                                & 0.823                                                       & 0.093              & 198.61                    & 0.536                                    & 0.527                                & 0.518        & 0.528                                    & 0.531                                  & 0.523                                      & 0.505                                           \\
      PPO            & --                      & 0.148                                & 0.825                                                       & 0.102              & 20.13                     & 0.636                                    & 0.532                                & 0.543        & 0.534                                    & 0.547                                  & 0.533                                      & 0.516                                           \\
      \midrule
      DP-SFT         & 1                       & 0.149                                & 0.831                                                       & 0.093              & 25.27                     & 0.554                                    & 0.522                                & 0.519        & 0.521                                    & 0.519                                  & 0.517                                      & 0.555                                           \\
      DP-SFT         & 3                       & 0.152                                & 0.831                                                       & 0.092              & 24.82                     & 0.534                                    & 0.521                                & 0.513        & 0.520                                    & 0.517                                  & 0.517                                      & 0.544                                           \\
      DP-SFT         & 5                       & 0.149                                & 0.830                                                       & 0.092              & 24.95                     & 0.544                                    & 0.519                                & 0.511        & 0.514                                    & 0.514                                  & 0.515                                      & 0.540                                           \\
      DP-SFT         & 7                       & \textbf{0.154}                       & \textbf{0.833}                                              & \textbf{0.094}     & \textbf{24.34}            & 0.550                                    & 0.520                                & 0.513        & 0.515                                    & 0.514                                  & 0.517                                      & 0.546                                           \\
      \midrule
      \sysname       & 1                       & 0.146                                & 0.839                                                       & 0.086              & 25.27                     & 0.552                                    & 0.522                                & 0.520        & 0.521                                    & 0.519                                  & 0.517                                      & 0.555                                           \\
      \sysname       & 3                       & 0.154                                & 0.838                                                       & 0.096              & 24.81                     & 0.536                                    & 0.521                                & 0.512        & 0.519                                    & 0.517                                  & 0.517                                      & 0.544                                           \\
      \sysname       & 5                       & 0.152                                & 0.837                                                       & 0.097              & 25.04                     & 0.546                                    & 0.519                                & 0.510        & 0.514                                    & 0.514                                  & 0.515                                      & 0.539                                           \\
      \sysname       & 7                       & \textbf{0.156}                       & 0.836                                                       & \textbf{0.103}     & 24.46                     & 0.546                                    & 0.520                                & 0.513        & 0.515                                    & 0.514                                  & 0.516                                      & 0.545                                           \\
      \bottomrule
    \end{tabular}%
  }
\end{table*}

\vspace{0.25em}
\textbf{DP accounting details.}
For each DP run, we record the dataset size $n$, sampling rate $q$, number of optimization steps $T$, clipping norm $C$, and noise multiplier $\sigma$ used by Opacus for both DP-SFT and DP reward modeling, and convert RDP~\cite{mironov2017renyi} to the reported $(\varepsilon_{\mathrm{SFT}},\delta)$ and $(\varepsilon_{\mathrm{RM}},\delta)$. We report the per-stage budgets and their composition
in Appendix~\ref{app:dp-accounting}.

\subsection{Experimental Setup}
\label{subsec:setup}

We evaluate \sysname on the held-out portion of test split, focusing on three desiderata: clinical utility, safety, and privacy protection. In all experimental conditions, we employ the \emph{base model} as the common backbone and maintain a fixed LoRA~\cite{hu2022lora} configuration to control for model capacity and optimization. This design allows us to isolate the effects of (i) the training objective (SFT vs.\ RLHF) and (ii) the privacy mechanism and associated privacy budget.

\vspace{0.25em}
\textbf{Model variants and baselines.}
Our evaluation considers five categories of models. First, we present results for the off-the-shelf \emph{base model} (without adaptation), which serves as a reference point for zero-adaptation performance. Next, we train non-private baselines using standard SFT and PPO-based~\cite{schulman2017proximal} RLHF without differential privacy; these models represent a practical upper bound on achievable utility in our training configuration. We also include a non-private SFT model trained for 20 epochs as an intentional \emph{Overfit SFT} baseline (PPL$\approx$198.6), designed to amplify memorization and provide a stress test for MIA sensitivity. 

To examine the utility--privacy trade-off under private learning, we train DP-SFT models with privacy budgets $\varepsilon \in \{1,3,5,7\}$ at fixed $\delta=10^{-5}$. Finally, we evaluate \sysname using our full DP-RLHF pipeline, in which a DP-SFT policy is aligned via PPO with a separately trained DP reward model. We match the cumulative privacy budget to the DP-SFT models (identical $\varepsilon$ values and $\delta$), allowing a direct assessment of whether the RLHF stage improves performance at equivalent privacy cost.

\subsubsection{Evaluation Metrics} \label{subsubsec:metrics}

We evaluate response quality using a set of complementary lexical, semantic, and domain-specific metrics calculated on the held-out test set. Unless otherwise specified, all metrics are derived by comparing the generated responses with the reference responses provided by physicians.

\paragraph{ROUGE-L} We compute ROUGE-L~\cite{lin2004rouge}, which measures the longest common subsequence overlap between the generated response and the reference. We report ROUGE-L (F1) with standard tokenization.

\paragraph{BERTScore} We compute \textit{BERTScore}(F1)~\cite{zhang2019bertscore}, a semantic similarity metric that uses contextual embeddings to align tokens between candidate and reference texts and combines precision and recall into an F1 score. We report the mean F1 value across the test set.

\paragraph{Perplexity (PPL)}
Perplexity is computed by conditioning on the prompt and evaluating only the reference response under teacher forcing, using the tokenizer of the model being assessed.

\paragraph{Medical entity F1} To account for clinically relevant information that may not be captured by surface-level overlap, we extract medical entities---such as conditions, medications, procedures, and anatomical terms---from both the generated and reference texts using a domain-specific named entity recognizer. We then compute the entity-level precision, recall, and F1. Unless otherwise specified, entity comparison is performed using exact string matching following normalization, which includes case folding and whitespace cleanup.

\subsubsection{Safety Evaluation} \label{subsubsec:safety}

To evaluate safety, we employ targeted heuristic checks aimed at identifying common failure modes in medical dialogue, including excessive diagnostic certainty, potentially harmful recommendations, and the omission of appropriate escalation for urgent symptoms. These checks are not a substitute for clinician evaluation, but they provide a consistent proxy for comparing model variants under matched prompts and decoding settings. We therefore interpret safety results as comparative signals within our experimental setup, and we discuss remaining limitations and failure cases in Section~\ref{sec:discussion}.

\subsubsection{Privacy Evaluation} \label{subsubsec:privacy}

We evaluate privacy leakage using two empirical attack protocols that capture complementary notions of memorization and membership inference. First, we conduct MIA using both loss-based signals and reference-based variants.
We report the Area Under the ROC Curve (AUC) and the true positive rate at 1\% false positive rate (TPR@1\%,FPR), which summarizes an attacker's ability to distinguish training members from non-members. Second, we measure verbatim memorization using canary exposure~\cite{carlini2021extracting}, in which synthetic canary strings are inserted into the training data and the trained model is tested for its ability to reproduce them under targeted prompting.

\section{Evaluation}
\label{subsec:evaluation}

We evaluate \sysname through four research questions aimed at assessing whether differentially private RLHF can produce a medical chatbot that remains clinically useful, safe, and privacy-preserving:
\circled{1} \emph{How much clinical utility does differential privacy cost, and can RLHF recover the gap?}
\circled{2} \emph{Does the DP training pipeline resist empirical privacy attacks relative to non-private fine-tuning?}
\circled{3} \emph{Does privacy-preserving training maintain or improve the safety profile of generated medical responses?}
\circled{4} \emph{At a fixed privacy budget, does preference alignment via RLHF deliver measurable gains over DP-SFT alone?}

Utility and privacy metrics appear in Table~\ref{tab:results}; safety and PubMedQA~\cite{jin2019pubmedqa} correctness results appear in Tables~\ref{tab:safety} and \ref{tab:correctness}, respectively. All metrics are computed on the held-out test set ($n=5{,}000$ for automatic utility metrics; $n=500$ for PubMedQA and safety). We report 1{,}000-iteration paired bootstrap 95\% confidence intervals where relevant.

\subsection{Utility--Privacy Trade-off}
\label{subsubsec:rq1-utility}

Table~\ref{tab:results} reveals a modest privacy-induced utility cost that RLHF largely mitigates. All adapted models reduce perplexity relative to the base model (PPL = 37.34). Non-private SFT achieves the lowest PPL (19.95), whereas DP-SFT and \sysname settle at 24.34--25.27. Despite the noise from DP-SGD~\cite{abadi2016deep}, lexical and semantic quality remain competitive: DP-SFT at $\varepsilon=7$ reaches ROUGE-L = 0.154 and BERTScore = 0.833 (Entity F1 = 0.094), while non-private SFT scores 0.163 / 0.829 / 0.096. Paired bootstrap tests confirm no statistically significant differences ($p>0.05$) between the strongest DP-SFT variant and non-private SFT on ROUGE-L or BERTScore, demonstrating that LoRA + DP-SGD~\cite{abadi2016deep} preserves core aspects of medical dialogue quality even at stringent privacy levels.

\begin{table}[h]
  \centering
  \setlength{\tabcolsep}{4.5pt} %
  \renewcommand{\arraystretch}{0.95} %
  \caption{PubMedQA~\cite{jin2019pubmedqa} correctness evaluation ($n{=}500$). Per-class accuracy shows the model's ability to distinguish ``yes,'' ``no,'' and ``maybe'' answers. Non-DP models achieve balanced predictions while DP models exhibit majority-class bias.}
  \label{tab:correctness}
  \footnotesize
  \begin{tabular}{llccccc}
    \toprule
    \textbf{Model} & $\boldsymbol{\varepsilon}$ & \textbf{Accuracy} & \textbf{Yes} & \textbf{No}    & \textbf{Maybe} & \textbf{Unans.} \\
    \midrule
    Base           & --                      & 0.554             & 1.000        & 0.006          & 0.000          & 0               \\
    SFT            & --                      & \textbf{0.642}    & 0.833        & \textbf{0.526} & 0.000          & 61              \\
    Overfit SFT    & --                      & 0.536             & 0.533        & 0.699          & 0.000          & 8               \\
    PPO            & --                      & 0.636             & 0.826        & 0.520          & 0.000          & 60              \\
    \midrule
    DP-SFT         & 1                       & 0.554             & 0.975        & 0.046          & 0.000          & 3               \\
    DP-SFT         & 3                       & 0.534             & 0.884        & 0.121          & 0.039          & 45              \\
    DP-SFT         & 5                       & 0.544             & 0.935        & 0.075          & 0.020          & 30              \\
    DP-SFT         & 7                       & 0.550             & 0.953        & 0.058          & 0.039          & 10              \\
    \midrule
    \sysname       & 1                       & 0.552             & 0.971        & 0.046          & 0.000          & 5               \\
    \sysname       & 3                       & 0.536             & 0.880        & 0.127          & 0.059          & 43              \\
    \sysname       & 5                       & 0.546             & 0.935        & 0.081          & 0.020          & 23              \\
    \sysname       & 7                       & 0.546             & 0.953        & 0.046          & 0.039          & 14              \\
    \bottomrule
  \end{tabular}
\end{table}

Utility does not scale monotonically with privacy budget. Across DP-SFT models ($\varepsilon\in\{1,3,5,7\}$), ROUGE-L varies narrowly between 0.149--0.154 and BERTScore between 0.830--0.833. The same pattern holds for \sysname, where the $\varepsilon=7$ variant attains the highest ROUGE-L (0.156) and Entity F1 (0.103) among all DP models. PubMedQA~\cite{jin2019pubmedqa} accuracy remains stable (0.534--0.554) but exhibits a majority-class bias toward ``yes'' answers (Table~\ref{tab:correctness}), a known side-effect of DP noise that warrants further mitigation in future work.

\begin{table}[h]
  \setlength{\tabcolsep}{2.5pt} %
  \renewcommand{\arraystretch}{0.95} %
\caption{LLM-Jury Evaluation Results (3-model G-EVAL~\cite{liu2023g} jury on 400 held-out test prompts). Higher scores are better.}
\label{tab:llm-jury}
\centering
\begin{tabular}{lccccccccc}
\toprule
Model & $\varepsilon$ & Fact. & Safety & Helpful. & Concise. & Empathy & Overall 
\\
\midrule
SFT & -- & 2.76 & 2.53 & 2.32 & 2.36 & 2.04 & 2.4 
\\
DP-SFT & 7 & 2.93 & 2.57 & 2.36 & 2.38 & 2.14 & 2.48 & 
\\
PrivMedChat & 7 & \textbf{3.18} & \textbf{3.1} & \textbf{2.62} & \textbf{2.99} & \textbf{2.41} & \textbf{2.86} 
\\
\bottomrule
\end{tabular}
\end{table}

Complementing these automatic and task-specific metrics, we conduct a holistic LLM-as-a-Judge evaluation using a 3-model G-EVAL~\cite{liu2023g} jury consisting of \texttt{Qwen/Qwen2.5-32B-Instruct}, \texttt{mistralai/Mistral-Small-24B-Instruct-2501}, and \texttt{google/gemma-2-27b-it} on 400 held-out test prompts (Table~\ref{tab:llm-jury}). Each response is scored on six clinically relevant dimensions: Factuality, Safety, Helpfulness, Conciseness, Empathy, and Overall quality. \sysname at $\varepsilon=7$ achieves the highest overall score (2.86), outperforming both non-private SFT (2.40) and DP-SFT (2.48). Gains are most pronounced in factuality, safety, and empathy, confirming that the DP-RLHF stage produces responses judged superior in overall clinical quality by strong LLM evaluators.

\subsection{Privacy Protection}
\label{subsubsec:rq2-privacy}

We assess empirical privacy via six MIAs following the SOFT~\cite{zhang2025soft} framework: loss~\cite{yeom2018privacy}, reference-model ratio, Min-K\%~\cite{shi2023detecting}, Min-K\%++~\cite{zhang2024min}, zlib~\cite{carlini2021extracting}, and lowercase~\cite{carlini2021extracting} normalization. As shown in Table~\ref{tab:results}, every DP model yields AUC-ROC values between 0.510--0.555---statistically indistinguishable from random guessing (0.50). The strongest signal (Min-K\% AUC = 0.521 for \sysname at $\varepsilon=1$) remains within the range expected under the null. Lowercase normalization consistently inverts the signal (AUC $< 0.50$), further confirming that members and non-members are indistinguishable.

Non-private baselines also exhibit near-chance MIA performance under our limited-epoch, LoRA regime; this does not imply general privacy for non-private fine-tuning but highlights that our training scale limits overfitting. DP models nevertheless provide the only formal $(\varepsilon,\delta=10^{-5})$ guarantee. Targeted canary extraction succeeded on zero of 25 inserted canaries across all models, indicating no detectable verbatim memorization.

\begin{table}[h]
  \centering
  \setlength{\tabcolsep}{4pt} %
  \renewcommand{\arraystretch}{0.95} %
  \caption{Safety evaluation results ($n{=}500$). All rates are reported as percentages (\%).}
  \label{tab:safety}
  \footnotesize
  \begin{tabular}{llcccc}
    \toprule
    \textbf{Model} & $\boldsymbol{\varepsilon}$ & \textbf{Halluc.} & \textbf{Harmful} & \textbf{Med.\ Valid.} & \textbf{Emerg.\ Esc.} \\
    \midrule
    Base           & --                      & 1.0              & 0.2              & 0.2                   & 5.0                   \\
    SFT            & --                      & 2.2              & 0.2              & 0.4                   & 4.6                   \\
    Overfit SFT    & --                      & 1.4              & 0.0              & 0.6                   & 3.8                   \\
    PPO            & --                      & 0.8              & 1.2              & 1.4                   & 4.8                   \\
    \midrule
    DP-SFT         & 1                       & 1.2              & 0.2              & 1.4                   & 3.8                   \\
    DP-SFT         & 3                       & 2.2              & 0.2              & 0.4                   & 5.0                   \\
    DP-SFT         & 5                       & 3.2              & 0.4              & 0.8                   & 4.6                   \\
    DP-SFT         & 7                       & 2.6              & 0.2              & 0.4                   & 3.8                   \\
    \midrule
    \sysname       & 1                       & 1.6              & 0.2              & 0.0                   & 4.4                   \\
    \sysname       & 3                       & 3.0              & 0.8              & 0.0                   & 4.2                   \\
    \sysname       & 5                       & 1.6              & 0.6              & 0.2                   & 3.8                   \\
    \sysname       & 7                       & 1.4              & 0.4              & 0.2                   & 4.4                   \\
    \bottomrule
  \end{tabular}
\end{table}

\subsection{Safety}
\label{subsubsec:rq3-safety}

Safety results (Table~\ref{tab:safety}) indicate that DP-RLHF improves safety margins beyond both non-private baselines and DP-SFT. Hallucination rates for DP-SFT range 1.2--3.2\%, exceeding non-private SFT (2.2\%). In contrast, \sysname achieves the lowest hallucination rates overall (1.4--3.0\%), with the $\varepsilon{=}7$ variant at 1.4\%. Harmful advice remains low across all \sysname variants (0.2--0.8\%), comparable to non-private baselines (0.2--1.2\%). Medication validation and emergency escalation rates stay low and stable ($\leq0.6\%$ and 3.8--5.0\%, respectively), showing that DP noise does not impair recognition of urgent clinical scenarios.

\subsection{RLHF Benefit under DP}
\label{subsubsec:rq4-rlhf}

At matched $\varepsilon$, \sysname (DP-RLHF) outperforms DP-SFT on multiple utility axes while preserving privacy. The $\varepsilon{=}7$ \sysname variant achieves the highest ROUGE-L (0.156) and Entity F1 (0.103) among all DP models. BERTScore remains uniformly strong (0.836--0.839). Safety gains are notable: mean hallucination drops from 2.3\% (DP-SFT) to 1.9\% (\sysname). Harmful advice rates remain near-zero across all models, with DP-SFT averaging 0.25\% and \sysname averaging 0.50\%---both well within confidence interval overlap. MIA AUC values for \sysname are statistically indistinguishable from DP-SFT at the same $\varepsilon$, consistent with DP-SGD~\cite{abadi2016deep} being applied across all training stages.

These results collectively demonstrate that DP-RLHF recovers most of the utility lost to privacy noise, strengthens safety alignment, and delivers formal privacy guarantees-advancing the feasibility of privacy-preserving medical LLMs.

\subsection{Ablation Studies}
\label{sec:ablation}

DP noise in the reward model does not degrade downstream policy quality; the resulting PPO policies match or exceed non-private PPO on ROUGE-L and BERTScore, underscoring the robustness of pairwise preference learning. Varying $\varepsilon$ within $\{1,3,5,7\}$ produces only narrow performance bands (ROUGE-L 0.146--0.156, BERTScore 0.836--0.839), allowing practitioners to prioritize stronger privacy with minimal utility penalty. The consistency of near-chance MIA performance across six attack families confirms that observed privacy is not attacker-specific.

\section{Discussion}
\label{sec:discussion}

Our findings establish DP-RLHF as a viable framework for training privacy-preserving medical dialogue models. \sysname maintains competitive automatic utility metrics, delivers measurable safety improvements over both non-private and DP-SFT baselines, and exhibits near-chance performance under a comprehensive suite of membership inference attacks. Among the evaluated privacy budgets, the $\varepsilon=7$ variant of \sysname provides the strongest overall operating point, achieving the highest ROUGE-L (0.156) and Entity F1 (0.103) among all DP models while preserving formal $(\varepsilon,\delta=10^{-5})$ guarantees and near-random MIA AUC values.

\vspace{0.25em}
\textbf{Privacy--Utility Trade-off.}
Utility metrics remain stable across privacy budgets ($\varepsilon\in\{1,3,5,7\}$), with ROUGE-L varying only between 0.146--0.156 and BERTScore between 0.836--0.839 for \sysname. This narrow performance band indicates that moderate DP noise, when combined with LoRA and preference alignment, acts as effective regularization without proportionally degrading downstream response quality. Crucially, increasing the privacy budget does not yield monotonic gains, suggesting that practitioners can select lower $\varepsilon$ values with minimal expected utility loss in this domain.

\vspace{0.25em}
\textbf{Safety and Clinical Relevance.}
DP-RLHF yields consistent safety gains. \sysname reduces hallucination rates to 1.4--3.0\% (versus 1.2--3.2\% for DP-SFT) and maintains low harmful advice rates across all budgets (0.2--0.8\%). These improvements demonstrate that the RLHF stage successfully internalizes preference-based safety constraints even under differential privacy. Medication validation and emergency escalation behaviors remain stable across all variants, indicating that DP noise does not impair critical clinical reasoning capabilities. Such enhancements are particularly relevant for clinician-assistive tools and patient-facing applications where factual accuracy and conservative risk management are paramount.

\vspace{0.25em}
\textbf{Privacy Guarantees.}
All DP models achieve AUC-ROC values between 0.510--0.555 across six distinct MIA families, statistically indistinguishable from random guessing. DP-SGD applied consistently across all training stages ensures comprehensive privacy protection, as confirmed by identical MIA performance between DP-SFT and \sysname at matched $\varepsilon$. Zero canary extractions across 25 inserted sequences further corroborate the absence of detectable verbatim memorization. While non-private baselines also show near-chance MIA signals under our limited-epoch regime, only DP models provide the formal $(\varepsilon,\delta)$ guarantee that bounds worst-case influence independent of attack strategy.

\vspace{0.25em}
\textbf{Limitations.}
Several limitations should be noted. First, DP-SGD introduces substantial computational overhead due to per-sample gradient clipping, noise addition, and multi-stage privacy accounting. Second, the construction of the preference data continues to depend on proxy reward models and a limited amount of expert annotation. Third, although our evaluation is extensive, it is primarily benchmark-driven and may not fully capture real-world clinical distribution shifts or integration into existing workflows. Fourth, our privacy analysis, while comprehensive, is inherently empirical and constrained by finite samples; the theoretical guarantees remain valid, but adaptive or future attacks may require ongoing examination

\vspace{0.25em}
\textbf{Ethical and Deployment Considerations.}
Differential privacy significantly mitigates the risk of memorization but does not provide a fully comprehensive solution. Responsible deployment requires complementary safeguards, including human oversight, continuous monitoring, explicit escalation protocols, institutional review board (IRB) approval where applicable, and governance frameworks. Formal privacy accounting should be regarded as one component of a broader framework encompassing safety, regulatory compliance, and clinical validation, rather than as a standalone certification.

\vspace{0.25em}
\textbf{Future Work.}
Promising research directions include stage-specific privacy budget allocation, the development of adaptive or more robust membership inference attacks, the extension to multimodal medical inputs, and prospective studies with clinician-in-the-loop evaluation. Improving uncertainty quantification and refusal mechanisms under DP noise remains a critical step toward safer real-world deployment of privacy-preserving medical LLMs.

\section{Related Work}
\label{sec:related}

We review prior work along three dimensions motivating \sysname: (i) medical dialogue systems and domain adaptation of large language models, (ii) preference-based alignment methods such as RLHF, and (iii) formal privacy mechanisms for language modeling. Existing medical chatbots focus on utility and safety but lack end-to-end privacy guarantees. In contrast, differentially private fine-tuning for LLMs rarely addresses preference optimization and reward modeling. This gap motivates our end-to-end DP-RLHF framework for medical dialogue alignment, which integrates scalable preference construction with rigorous privacy accounting and empirical leakage evaluation.

\subsection{Medical Dialogue Systems and LLM Fine-Tuning}

The development of medical dialogue systems has advanced significantly with the adaptation of LLMs to clinical applications. Initial research centered on task-specific models trained on doctor-patient conversation datasets, such as MedDialog \cite{chen2020meddialog}, which provides large-scale de-identified dialogues for supervised fine-tuning. Subsequent studies, including MedAlpaca \cite{han2023medalpaca} and ChatDoctor \cite{li2023chatdoctor}, demonstrated that domain-adaptive pre-training or instruction fine-tuning on medical corpora---such as PubMed abstracts, clinical notes, and online consultations---substantially improves diagnostic reasoning and response quality. More recent models, including BioMistral \cite{labrak2024biomistral} and OpenBioLLMs \cite{OpenBioLLMs}, build upon open-source backbones (Mistral, Llama-3) fine-tuned on biomedical literature and synthetic dialogues, achieving strong results on benchmarks such as MedQA \cite{jin2020disease} and HealthSearchQA \cite{singhalLargeLanguageModels2023}.

These systems prioritize personalization by conditioning responses on patient history and symptoms. However, they depend heavily on SFT or instruction tuning on potentially sensitive data, which raises significant privacy concerns that remain largely unresolved.

\subsection{RLHF in Healthcare}

RLHF has become a central approach in aligning LLMs, enabling them to generate outputs that are helpful, safe, and aligned with user preferences \cite{ouyang2022training, christiano2017deep}. In general domains, RLHF pipelines---comprising reward modeling from human preferences and policy optimization via PPO~\cite{schulman2017proximal} or Direct Policy Optimization (DPO)~\cite{rafailov2023direct}---have underpinned systems such as InstructGPT \cite{ouyang2022training} and Llama-3 \cite{dubey2024llama}. 

In the healthcare domain, however, RLHF adoption remains limited due to substantial costs and ethical complexities associated with collecting clinician preferences data. Proposed alternatives include reward modeling from synthetic preferences \cite{liu2025skywork} or rule-based signals, as demonstrated by ClinicalCamel \cite{toma2023clinical}, as well as Reinforcement Learning from AI Feedback (RLAIF) methods that employ LLM-as-judge for ranking \cite{bai2022constitutional}. Efforts like PMC-LLaMA \cite{wu2024pmc}, and CareBot \cite{zhao2025carebot} optimize medical instruction datasets to improve safety and empathy. However, none provide formal privacy guarantees during preference-based alignment.

\subsection{Differential Privacy in Language Models}

DP defends against memorization and inference attacks in machine learning \cite{dwork2014algorithmic}. In LLMs, DP-SGD~\cite{abadi2016deep} has been successfully applied in pre-training \cite{abadi2016deep} and fine-tuning \cite{yu2021differentially}, with ghost clipping and parameter-efficient methods (e.g., LoRA~\cite{hu2022lora}) mitigating utility loss \cite{liu2025differentially}. Recent advances explore DP at various granularities---such as example-level, user-level---and stages, showing viable privacy-utility trade-offs in general text tasks \cite{charles2024fine, li2024fine}. However, extending DP to the full RLHF pipeline---particularly reward modeling and PPO~\cite{schulman2017proximal}/DPO~\cite{rafailov2023direct}---remains underexplored due to heightened noise sensitivity in preference optimization \cite{wu2023privately}. To our knowledge, no prior work has applied end-to-end DP-RLHF in the medical domain.

\subsection{Privacy Attacks and Defenses in Health Informatics}

LLMs trained on private data are vulnerable to MIA \cite{shokri2017membership, carlini2021extracting}, attribute inference, and training data extraction \cite{carlini2021extracting}, with demonstrated risks in biomedical text \cite{lehman2021does}. In health informatics, these threats are amplified by regulatory requirements (HIPAA, GDPR) and the sensitivity of protected health information \cite{nasr2023scalable}. Defenses beyond DP include redaction, synthetic data generation \cite{halterman2025synthetically}, and auditing frameworks \cite{steinke2023privacy}, but empirical evaluations in medical dialogue systems are sparse. Benchmarks show substantial memorization in domain-adapted medical LLMs \cite{singarajah2023benchmarking}, highlighting the need for rigorous privacy during fine-tuning and alignment.

Our work bridges these areas by introducing the first differentially private RLHF framework tailored for personalized medical dialogue, with scalable annotation-free preference construction and comprehensive clinical evaluation.

\section{Conclusion}
\label{sec:conclusion}

We introduced \sysname, an end-to-end differentially private RLHF framework for privacy-preserving medical dialogue alignment. \sysname applies DP-SGD to supervised fine-tuning, reward modeling, and PPO policy optimization, providing formal $(\varepsilon,\delta)$ guarantees with competitive utility and enhanced safety. On the held-out test set, the $\varepsilon{=}7$ variant achieves the best privacy-utility-safety balance among DP models, with the highest ROUGE-L and Entity F1 scores, reduced hallucinations, near-zero harmful advice, and robust empirical privacy, including near-chance performance in six membership inference attacks and zero canary extractions. These results show that privacy-preserving alignment of clinical LLMs is practical and effective, recovering most utility lost to DP noise while strengthening safety-critical behaviors. The framework is readily extensible to private preference optimization in other high-stakes domains.

\section*{Acknowledgment}
We would thank the University of Kentucky Center for Computational Sciences and Information Technology Services Research Computing for their support and use of the Lipscomb Compute Cluster and associated research computing resources.

\bibliographystyle{IEEEtran}
\bibliography{references}

\begin{appendices}

  \section{Preliminaries}
  \label{sec:prelim}

  This section reviews the privacy and alignment concepts underlying \sysname.

  \subsection{Differential Privacy}
  \label{subsec:dp}

  Let $D = \{z_i\}_{i=1}^n$ denote a dataset of $n$ records. In our setting, each record $z_i$ corresponds to a training example derived from a doctor--patient dialogue (e.g., a single supervised input--output pair or a single preference pair). Two datasets $D$ and $D'$ are \emph{adjacent} if they differ in exactly one record. Unless stated otherwise, we consider \emph{example-level} adjacency; we discuss conversation-level adjacency in Sec.~\ref{subsec:dp-adjacency}.

  \begin{definition}[($\varepsilon$, $\delta$)-Differential Privacy~\cite{dwork2014algorithmic}]
    A randomized mechanism $\mathcal{M}: \mathcal{D}\rightarrow \mathcal{R}$ satisfies $(\varepsilon,\delta)$-DP if for all adjacent datasets $D,D'$ and all measurable sets $S\subseteq \mathcal{R}$,
    \[
      \Pr[\mathcal{M}(D)\in S] \le e^{\varepsilon}\Pr[\mathcal{M}(D')\in S]+\delta.
    \]
  \end{definition}

  Intuitively, $(\varepsilon,\delta)$-DP upper-bounds how much the presence of any single record can change the distribution of the mechanism's output. Smaller $\varepsilon$ implies stronger privacy; $\delta$ is typically set to a negligible value (e.g., $\delta=10^{-5}$ in our experiments).

  \subsubsection{DP-SGD for neural network training}
  \label{subsec:dpsgd}

  We train our models with differential privacy using DP-SGD~\cite{abadi2016deep}. In each iteration, DP-SGD~\cite{abadi2016deep} computes per-example gradients, clips each gradient to an $\ell_2$ norm bound (clipping norm $C$) to control sensitivity, then adds Gaussian noise proportional to $C$ (scaled by the noise multiplier $\sigma$) to the aggregated minibatch gradient before applying the update. We use Opacus~\cite{yousefpour2021opacus} for the DP-SGD~\cite{abadi2016deep} implementation and privacy accounting, and report $(\varepsilon,\delta)$ using an RDP accountant~\cite{mironov2017renyi}.

  \subsubsection{Adjacency choices and clinical data granularity}
  \label{subsec:dp-adjacency}

  While our default definition uses example-level adjacency, healthcare deployments may demand stronger notions such as \emph{conversation-level} or \emph{user-level} DP, where adjacency corresponds to adding/removing all examples from a single patient or a single dialogue session. Formally, let $D=\bigcup_{u=1}^m D_u$ be a partition of records by user/session $u$. User-level adjacency treats $D$ and $D'$ as adjacent if one entire group $D_u$ is added or removed. This setting typically increases effective sensitivity and thus requires either larger noise or fewer training steps to achieve the same $\varepsilon$. In \sysname, we mitigate cross-session leakage in evaluation by split construction (group-aware partitioning) and reserve user-level accounting as a direct extension when deployment requirements demand it.

  \subsection{Reinforcement Learning from Human Feedback}
  \label{subsec:rlhf}

  RLHF~\cite{ouyang2022training} aligns a generative policy $\pi_\theta(y\mid x)$ with human or clinical preferences. Let $x$ denote an input prompt (e.g., patient message plus context) and $y$ a candidate response. Standard RLHF~\cite{ouyang2022training} proceeds via (i) supervised fine-tuning (SFT), (ii) reward modeling from preference pairs, and (iii) policy optimization with a KL-regularized RL objective.

  \subsubsection{Supervised fine-tuning}
  SFT initializes the policy by maximizing the conditional likelihood of reference responses:
  \[
    \max_{\theta}\ \mathbb{E}_{(x,y)\sim D_{\text{SFT}}}\left[\log \pi_\theta(y\mid x)\right].
  \]
  In medical dialogue, SFT primarily teaches domain style, common clinical reasoning patterns, and appropriate tone.

  \subsubsection{Preference modeling and reward learning}
  \label{subsec:reward-model}

  Reward modeling uses preference pairs $(x, y_w, y_l)$ where $y_w$ is preferred over $y_l$ (``winner'' vs.\ ``loser''). A reward model $r_\phi(x,y)\in\mathbb{R}$ is trained to satisfy $r_\phi(x,y_w) > r_\phi(x,y_l)$. A common objective is the logistic (Bradley--Terry) loss:
  \[
    \mathcal{L}_{\text{RM}}(\phi)=
    -\mathbb{E}_{(x,y_w,y_l)}\left[\log \sigma\left(r_\phi(x,y_w)-r_\phi(x,y_l)\right)\right],
  \]
  where $\sigma(\cdot)$ is the sigmoid. The learned reward model induces an ordering over responses that approximates clinical preferences (e.g., safety, correctness, empathy, appropriate escalation).

  \subsubsection{Policy optimization with PPO}
  \label{subsec:ppo}

  Given a fixed reward model $r_\phi$, we optimize the policy to maximize expected reward while constraining deviation from a reference policy $\pi_{\text{ref}}$ (typically the SFT model). This constraint is critical in medical settings to prevent reward hacking and regressions in safety.

  A standard objective is the KL-regularized expected reward:

  {
  \footnotesize
  \begin{equation*}
    \begin{aligned}
      \max_\theta\ \mathbb{E}_{x\sim \mathcal{X},\, y\sim \pi_\theta(\cdot\mid x)}
      \left[r_\phi(x,y)\right]
      -\beta\ \mathbb{E}_{x}\left[\mathrm{KL}\big(\pi_\theta(\cdot\mid x)\ \|\ \pi_{\text{ref}}(\cdot\mid x)\big)\right],
    \end{aligned}
  \end{equation*}
  }

  where $\beta>0$ controls the strength of regularization.

  In practice, we use Proximal Policy Optimization (PPO)~\cite{schulman2017proximal}. Let $\rho_t(\theta)=\frac{\pi_\theta(a_t\mid s_t)}{\pi_{\theta_{\text{old}}}(a_t\mid s_t)}$ denote the policy ratio at token step $t$, and $\hat{A}_t$ an advantage estimate computed from reward and (optionally) a learned value function. PPO~\cite{schulman2017proximal} maximizes the clipped surrogate:

  {
  \footnotesize
  \begin{equation*}
    \begin{aligned}
      \mathcal{L}^{\text{PPO}}(\theta)
      =\mathbb{E}\Big[
        \min\Big(
        \rho_t(\theta)\hat{A}_t,
        \text{clip}\big(\rho_t(\theta),\,1-\varepsilon_{\text{clip}},\,1+\varepsilon_{\text{clip}}\big)\hat{A}_t
        \Big)
        \Big].
    \end{aligned}
  \end{equation*}
  }

  often augmented with a KL penalty to $\pi_{\text{ref}}$ and an entropy bonus to maintain exploration.

  \subsection{Parameter-Efficient Fine-Tuning}
  \label{subsec:lora}

  Large medical LLMs are expensive to fine-tune under DP because DP-SGD~\cite{abadi2016deep} requires per-example gradients and introduces noise. We therefore adopt Low-Rank Adaptation (LoRA)~\cite{hu2022lora} to reduce the number of trainable parameters while preserving expressivity.

  For a linear projection with weight matrix $W\in \mathbb{R}^{d_{\text{out}}\times d_{\text{in}}}$, LoRA re-parameterizes updates as
  \[
    W' = W + \Delta W,\qquad \Delta W = B A,
  \]
  where $A\in\mathbb{R}^{r\times d_{\text{in}}}$, $B\in\mathbb{R}^{d_{\text{out}}\times r}$, and $r\ll \min(d_{\text{in}},d_{\text{out}})$ is the rank. The pretrained $W$ is frozen and only $(A,B)$ are trained.

  \subsubsection{LoRA under DP-SGD}
  Training only LoRA~\cite{hu2022lora} parameters reduces (i) memory footprint, (ii) per-step compute for per-example gradients, and (iii) the dimensionality of the parameter space into which DP noise is injected. While DP guarantees are independent of parameter count, in practice LoRA~\cite{hu2022lora} can improve the privacy--utility trade-off by enabling larger effective batch sizes and more stable optimization under clipping and noise.

  \section{DP Accounting Details}
  \label{app:dp-accounting}
  We report the full DP accounting configuration for each stage in Table~\ref{tab:dp_accounting}, including dataset size $n$, sampling rate $q$, steps $T$, clipping norm $C$, noise multiplier $\sigma$, and the resulting $(\varepsilon,\delta)$ after RDP~\cite{mironov2017renyi} conversion. This appendix also specifies the allocation $(\varepsilon_{\mathrm{SFT}},\varepsilon_{\mathrm{RM}},\varepsilon_{\mathrm{PPO}})$ used for each target $\varepsilon_{\mathrm{total}}$.

  \begin{table*}[h]
    \centering
    \caption{Differential Privacy Accounting Parameters per Stage. Total budget $\varepsilon_{\text{total}}$ is obtained via composition over the three stages at fixed $\delta=10^{-5}$.}
    \label{tab:dp_accounting}
    \small
    \begin{tabular}{llrrrrrrr}
      \toprule
      Budget & Stage & $n$     & $q$                    & $T$     & $C$ & $\sigma$ & $\varepsilon$ & $\delta$  \\
      \midrule
      \multirow{3}{*}{$\varepsilon_{\text{total}}{=}1$}
             & SFT   & 174,957 & $2.2863\times 10^{-5}$ & 131,217 & 1.0 & 0.5640   & 1.00          & $10^{-5}$ \\
             & RM    & 27,410  & $7.2966\times 10^{-5}$ & 41,115  & 1.0 & 0.7642   & 1.00          & $10^{-5}$ \\
             & PPO   & 25,968  & $3.9063\times 10^{-3}$ & 7,680   & 1.0 & 3.7891   & 1.00          & $10^{-5}$ \\
      \midrule
      \multirow{3}{*}{$\varepsilon_{\text{total}}{=}3$}
             & SFT   & 174,957 & $2.2863\times 10^{-5}$ & 131,217 & 1.0 & 0.4660   & 3.00          & $10^{-5}$ \\
             & RM    & 27,410  & $7.2966\times 10^{-5}$ & 41,115  & 1.0 & 0.5847   & 3.00          & $10^{-5}$ \\
             & PPO   & 25,968  & $3.9063\times 10^{-3}$ & 7,680   & 1.0 & 1.5625   & 3.00          & $10^{-5}$ \\
      \midrule
      \multirow{3}{*}{$\varepsilon_{\text{total}}{=}5$}
             & SFT   & 174,957 & $2.2863\times 10^{-5}$ & 131,217 & 1.0 & 0.4134   & 5.00          & $10^{-5}$ \\
             & RM    & 27,410  & $7.2966\times 10^{-5}$ & 41,115  & 1.0 & 0.5072   & 5.00          & $10^{-5}$ \\
             & PPO   & 25,968  & $3.9063\times 10^{-3}$ & 7,680   & 1.0 & 1.1316   & 5.00          & $10^{-5}$ \\
      \midrule
      \multirow{3}{*}{$\varepsilon_{\text{total}}{=}7$}
             & SFT   & 174,957 & $2.2863\times 10^{-5}$ & 131,217 & 1.0 & 0.3793   & 7.00          & $10^{-5}$ \\
             & RM    & 27,410  & $7.2966\times 10^{-5}$ & 41,115  & 1.0 & 0.4547   & 7.00          & $10^{-5}$ \\
             & PPO   & 25,968  & $3.9063\times 10^{-3}$ & 7,680   & 1.0 & 0.9485   & 7.00          & $10^{-5}$ \\
      \bottomrule
    \end{tabular}
  \end{table*}

  \section{Qualitative Generation Examples}
  \label{app:qualitative-examples}

  To complement aggregate metrics, we report representative generations from the best-performing DP model in Table~\ref{tab:results}, i.e., \sysname at $\varepsilon{=}7$.
  We also show responses from the base (non-finetuned) model to illustrate the effect of DP-RLHF alignment.
  Examples are drawn from the held-out test predictions.

  \noindent\textbf{Case A (Cardiology / Critical Care).}\ \textit{Prognosis after cardiac arrest and induced coma}

  \begin{tcolorbox}[
      colback=queryblue, colframe=queryblue!60!black,
      title={\textbf{Patient Query}},
      fonttitle=\bfseries\small, fontupper=\small,
      boxrule=0.4pt, arc=2pt, left=4pt, right=4pt, top=2pt, bottom=2pt
    ]
    My father suffered a heart attack (male, normal weight) and his heart stopped beating. He was placed in a medical induced coma. He had 3 prior heart attacks in the past. The doctors have taken him off the medication to bring him out of the coma. His eyes are open. I noticed his finger twitching. He is breathing on a machine. Can you predict a prognosis?
  \end{tcolorbox}

  \begin{tcolorbox}[
      colback=baseorange, colframe=baseorange!60!black,
      title={\textbf{Base Model Response (no fine-tuning)}},
      fonttitle=\bfseries\small, fontupper=\small,
      boxrule=0.4pt, arc=2pt, left=4pt, right=4pt, top=2pt, bottom=2pt
    ]
    I'm so sorry to hear about your father's condition. Given his history of heart attacks and the severity of his current condition, it's essential to understand that his prognosis is uncertain. However, I can provide you with some general information about what you can expect during his recovery.

    Firstly, it's great to hear that he's coming out of the induced coma, and his eyes are open. This is a positive sign, indicating that his brain is starting to recover.

    The twitching you've observed is likely due to the sedatives and medications used to keep him comfortable during his coma. As he continues to recover, these medications will...
  \end{tcolorbox}

  \begin{tcolorbox}[
      colback=modelgreen, colframe=modelgreen!60!black,
      title={\textbf{\sysname Response ($\varepsilon{=}7$)}},
      fonttitle=\bfseries\small, fontupper=\small,
      boxrule=0.4pt, arc=2pt, left=4pt, right=4pt, top=2pt, bottom=2pt
    ]
    Hello,

    Thank you for your query. I have reviewed your query and understand your concerns.

    Your father has had a cardiac arrest and has been in a coma. He has had 3 prior heart attacks and has significant heart damage. He is now coming out of the coma and is breathing on his own with the help of a machine.

    The prognosis is poor. He may not survive. He may have significant brain damage and may not be able to recover. He may need to be in the intensive care unit for a long time. He may need to be on a ventilator for a long time.
  \end{tcolorbox}

  \begin{tcolorbox}[
      colback=refgray, colframe=refgray!60!black,
      title={\textbf{Reference Physician Response}},
      fonttitle=\bfseries\small, fontupper=\small,
      boxrule=0.4pt, arc=2pt, left=4pt, right=4pt, top=2pt, bottom=2pt
    ]
    Hi there, see to be very true prognosis is poor. But never stop believing in hope because miracles happen. As per the information provided, your father has major heart attack. After a single heart attack, heart muscles got damage and start recovering very slowly after revascularisation. Considering your father has three MI and he is on ventilator for long time, he is going to recover slowly. But life expectancy will not be more. Wish him Luck. Good Luck.
  \end{tcolorbox}

  \vspace{4mm}

  \noindent\textbf{Case B (Oncology).}\ \textit{Newly diagnosed breast cancer without symptoms}

  \begin{tcolorbox}[
      colback=queryblue, colframe=queryblue!60!black,
      title={\textbf{Patient Query}},
      fonttitle=\bfseries\small, fontupper=\small,
      boxrule=0.4pt, arc=2pt, left=4pt, right=4pt, top=2pt, bottom=2pt
    ]
    I was not having pain but still I have come to know that I have breast cancer. But I never had any problem. What do you think, should I go for other opinion? Should I consult some other doctor?
  \end{tcolorbox}

  \begin{tcolorbox}[
      colback=baseorange, colframe=baseorange!60!black,
      title={\textbf{Base Model Response (no fine-tuning)}},
      fonttitle=\bfseries\small, fontupper=\small,
      boxrule=0.4pt, arc=2pt, left=4pt, right=4pt, top=2pt, bottom=2pt
    ]
    Hi there, I understand your concern and it's completely normal to feel overwhelmed when you receive a diagnosis of breast cancer, especially if you didn't have any symptoms. First of all, congratulations on taking the first step by seeking medical attention. It's great that you're considering getting a second opinion.

    In general, it's a good idea to get a second opinion, especially when it comes to a diagnosis like breast cancer. A second opinion can help you feel more confident in your treatment plan and ensure that you're getting the best possible care.

    Here are a few things you can do:

    1. Ask your primary care physician or the doctor...
  \end{tcolorbox}

  \begin{tcolorbox}[
      colback=modelgreen, colframe=modelgreen!60!black,
      title={\textbf{\sysname Response ($\varepsilon{=}7$)}},
      fonttitle=\bfseries\small, fontupper=\small,
      boxrule=0.4pt, arc=2pt, left=4pt, right=4pt, top=2pt, bottom=2pt
    ]
    Hello,

    I understand your concern. I would like to reassure you that breast cancer is a treatable condition. It is not necessary to have symptoms to have breast cancer. Many women are diagnosed with breast cancer without any symptoms.

    If you have been diagnosed with breast cancer, it is essential to follow the treatment plan recommended by your doctor. It is not necessary to consult another doctor unless you have any concerns about your treatment plan.

    In addition to your treatment plan, you can also consider the following:

    1. Follow a healthy diet: Eat a balanced diet rich in fruits, vegetables, and whole grains.
    2. Exercise regularly: Engage in...
  \end{tcolorbox}

  \begin{tcolorbox}[
      colback=refgray, colframe=refgray!60!black,
      title={\textbf{Reference Physician Response}},
      fonttitle=\bfseries\small, fontupper=\small,
      boxrule=0.4pt, arc=2pt, left=4pt, right=4pt, top=2pt, bottom=2pt
    ]
    A lump or thickening in or near the breast, nipple tenderness. A change in the size or shape of the breast. A nipple turned inward into the breast. The skin of the breast, areola, or nipple may be scaly, red, or swollen. It may have ridges or pitting so that it looks like the skin of an orange. Nipple discharge (fluid). They conducted some special classes in our office regarding breast cancer... that is how I know about this breast cancer.
  \end{tcolorbox}

  \vspace{4mm}

  \noindent\textbf{Case C (Orthopedics / Pain Management).}\ \textit{Disc herniation with severe leg pain}

  \begin{tcolorbox}[
      colback=queryblue, colframe=queryblue!60!black,
      title={\textbf{Patient Query}},
      fonttitle=\bfseries\small, fontupper=\small,
      boxrule=0.4pt, arc=2pt, left=4pt, right=4pt, top=2pt, bottom=2pt
    ]
    I have been diagnosed with a disc hernia through MRI L4L5. I have an apt. to see an ortho Dr. on the 27th but have developed new symptoms. Please advice: Fever, body aches and stiffness in my back, numbness down my left leg accompanied by extreme pain. I am crying as I am in so much pain. What should I do?
  \end{tcolorbox}

  \begin{tcolorbox}[
      colback=baseorange, colframe=baseorange!60!black,
      title={\textbf{Base Model Response (no fine-tuning)}},
      fonttitle=\bfseries\small, fontupper=\small,
      boxrule=0.4pt, arc=2pt, left=4pt, right=4pt, top=2pt, bottom=2pt
    ]
    I'm so sorry to hear that you're experiencing such severe symptoms. It's understandable that you're worried and in pain. I'll do my best to provide you with some guidance and advice until you see your orthopedic doctor on the 27th.

    Firstly, it's essential to understand that a disc hernia can cause a range of symptoms, including pain, numbness, and stiffness. The pain and numbness you're experiencing in your left leg are likely due to the compression of the nerve roots in your lower back.

    To manage your symptoms, I recommend the following:

    1. \textbf{Pain relief medication}: You can...
  \end{tcolorbox}

  \begin{tcolorbox}[
      colback=modelgreen, colframe=modelgreen!60!black,
      title={\textbf{\sysname Response ($\varepsilon{=}7$)}},
      fonttitle=\bfseries\small, fontupper=\small,
      boxrule=0.4pt, arc=2pt, left=4pt, right=4pt, top=2pt, bottom=2pt
    ]
    Hi,

    I understand your concern. You should take pain relief medication like acetaminophen or ibuprofen. You should also take muscle relaxants like cyclobenzaprine or carisoprodol. You should also take anti-inflammatory medication like naproxen or indomethacin.

    You should also take a warm bath or use a heating pad to relieve the pain. You should also avoid heavy lifting and bending.
  \end{tcolorbox}

  \begin{tcolorbox}[
      colback=refgray, colframe=refgray!60!black,
      title={\textbf{Reference Physician Response}},
      fonttitle=\bfseries\small, fontupper=\small,
      boxrule=0.4pt, arc=2pt, left=4pt, right=4pt, top=2pt, bottom=2pt
    ]
    Hi, your symptoms are due to pinched nerve in the lower back region. For relief of pain, you can take pregabalin or gabapentin capsules. Also, you should be on complete bed rest. If pain does not subside, you may benefit from taking epidural injections.
  \end{tcolorbox}

  \vspace{4mm}

  \noindent\textbf{Case D (Allergy / Immunology).}\ \textit{Chronic hives from dust allergy}

  \begin{tcolorbox}[
      colback=queryblue, colframe=queryblue!60!black,
      title={\textbf{Patient Query}},
      fonttitle=\bfseries\small, fontupper=\small,
      boxrule=0.4pt, arc=2pt, left=4pt, right=4pt, top=2pt, bottom=2pt
    ]
    I have a friend who has been suffering with hives for approximately two years, and her eyes get real red and she just doesn't feel good when these hives come on. She had allergy testing done and is barely allergic to dust/dust mites, and has cleaned home from top to bottom. What do you think?
  \end{tcolorbox}

  \begin{tcolorbox}[
      colback=baseorange, colframe=baseorange!60!black,
      title={\textbf{Base Model Response (no fine-tuning)}},
      fonttitle=\bfseries\small, fontupper=\small,
      boxrule=0.4pt, arc=2pt, left=4pt, right=4pt, top=2pt, bottom=2pt
    ]
    Hi there! I'm happy to help you with your friend's concerns. Hives can be a frustrating and uncomfortable symptom, especially when they're triggered by an allergy. Based on your friend's diagnosis of a dust allergy, I'll provide some suggestions for treatment and management.

    \textbf{Treatment Options:}

    1. \textbf{Antihistamines:} Over-the-counter (OTC) or prescription antihistamines can help relieve itching, redness, and swelling associated with hives. Examples include diphenhydramine (Benadryl), loratadine (Claritin), and cetirizine (Zyrtec).
  \end{tcolorbox}

  \begin{tcolorbox}[
      colback=modelgreen, colframe=modelgreen!60!black,
      title={\textbf{\sysname Response ($\varepsilon{=}7$)}},
      fonttitle=\bfseries\small, fontupper=\small,
      boxrule=0.4pt, arc=2pt, left=4pt, right=4pt, top=2pt, bottom=2pt
    ]
    Hi,

    I understand your concern. Hives are a common symptom of allergy. In your friend's case, it is likely that the hives are caused by dust mites. Here are some suggestions:

    1. Use a HEPA air purifier in the home to remove dust mites from the air.
    2. Use a HEPA vacuum cleaner to remove dust mites from the carpets and furniture.
    3. Wash all bedding and clothing in hot water and dry them in a hot dryer.
    4. Use a dehumidifier to reduce the moisture in the home.
    5. Use a dust mite spray on the mattress and pillows.
  \end{tcolorbox}

  \begin{tcolorbox}[
      colback=refgray, colframe=refgray!60!black,
      title={\textbf{Reference Physician Response}},
      fonttitle=\bfseries\small, fontupper=\small,
      boxrule=0.4pt, arc=2pt, left=4pt, right=4pt, top=2pt, bottom=2pt
    ]
    Allergy is a challenge for patient and sometimes doctor too. Please observe a few precautions: if you are allergic to house dust please cover your nose with handkerchief when you enter your home from outside visit. Regular dusting of carpets, curtains, bed covers, sofa covers, and keep your pet away. Wet mopping of floor etc. or vacuum cleaning to be done. As a medical part, keep some antiallergic at hand at all times as you may need it any time, and please rush to the ER if you happen to get extensive allergy. If possible visit an allergist and if possible get desensitization done.
  \end{tcolorbox}

  \vspace{4mm}

  \noindent\textbf{Observation.}
  Across all four cases, \sysname ($\varepsilon{=}7$) generates more direct, clinically structured responses compared to the base model. The base model typically begins with empathetic but non-specific phrasing (``I'm so sorry to hear...'', ``Hi there! I'm happy to help...'') and often defers with follow-up questions rather than providing actionable guidance. In contrast, \sysname offers concrete recommendations---specific medication names (Case~C), environmental control measures (Case~D), realistic prognosis assessment (Case~A), and reassurance with practical advice (Case~B)---more closely mirroring the reference physician responses. Notably, in Case~A, \sysname correctly conveys a guarded prognosis consistent with the physician reference, whereas the base model provides overly optimistic framing.

\end{appendices}

\end{document}